\documentclass[letterpaper, 10 pt, conference]{ieeeconf}  

\IEEEoverridecommandlockouts                              
\makeatletter

\let\proof\@undefined
\let\endproof\@undefined
\let\IEEElabelindent\labelindent
\makeatother

\usepackage{graphicx}
\usepackage{caption}
\usepackage{subcaption}
\usepackage{enumerate}
\usepackage{hyperref}
\usepackage{upgreek}
\usepackage{mathrsfs}
\usepackage[table]{xcolor}
\usepackage{amsmath} 
\usepackage{amssymb}  
\usepackage{amsthm}  
\usepackage{bm}
\usepackage{lipsum}
\usepackage{color}
\usepackage{multirow, array}
\usepackage{cite}
\usepackage[linesnumbered, ruled, vlined]{algorithm2e}

\newcommand{\saa}{ \mbox{\tiny SAA}}

\title{\LARGE \bf
Infusing Model Predictive Control into Meta-Reinforcement Learning 
for Mobile Robots in Dynamic Environments
\thanks{This work was supported in part by  the National Research Foundation of Korea funded by MSIT(2020R1C1C1009766, 2021R1A4A2001824), the Information and Communications Technology Planning and Evaluation (IITP) grant funded by MSIT(2020-0-00857, 2022-0-00480), and Samsung Electronics.}
}

\author{Jaeuk Shin \and
Astghik Hakobyan \and
Mingyu Park \and
Yeoneung Kim \and
Gihun Kim \and
Insoon Yang 
\thanks{
All authors are with the Department of Electrical and Computer Engineering, Automation and Systems Research Institute,  Seoul National University, Seoul 08826, South Korea, 
        {\tt\small insoonyang@snu.ac.kr}}%
}

\begin{document}

\maketitle
\thispagestyle{empty}
\pagestyle{empty}


\begin{abstract}
The successful operation of mobile robots requires them to adapt rapidly to environmental changes.
To develop an adaptive decision-making tool for mobile robots, we propose a novel algorithm that combines meta-reinforcement learning (meta-RL) with model predictive control (MPC). 
Our method employs an off-policy meta-RL algorithm as a baseline to train a policy using transition samples generated by MPC when the robot detects certain events that can be effectively handled by MPC, with its explicit use of robot dynamics. 
The key idea of our method is to switch between the meta-learned policy and the MPC controller in a randomized and event-triggered fashion
to make up for suboptimal MPC actions caused by the limited prediction horizon. 
During meta-testing, the MPC module is deactivated to significantly reduce computation time in  motion control. 
We further propose an online adaptation scheme that enables the robot to  infer and adapt to a new task within a single trajectory.
The performance of our method has been demonstrated through 
simulations using a nonlinear car-like vehicle model with $(i)$ synthetic movements of obstacles, and $(ii)$ real-world pedestrian motion data. 
The simulation results indicate that our method outperforms other algorithms in terms of learning efficiency and navigation quality.
\end{abstract}


\section{Introduction}\label{sec:intro}

As mobile robots operate in practical environments cluttered with moving obstacles,
it is crucial for them to  adapt quickly to environmental changes. 
For example, a service robot in a restaurant should be able to  move efficiently and safely to a goal position
regardless of changes in the motion pattern of other service robots and human agents as well as the configuration of tables. 
The focus of this work is to develop an adaptive sequential decision-making tool 
that combines meta-reinforcement learning (meta-RL) and model predictive control (MPC) to attain the advantages of these two complementary methods.

Reinforcement learning (RL) algorithms are popular tools for solving robot navigation problems~(e.g., \cite{faust2018prm}). However, RL algorithms are unable to generalize the learned policy to new tasks or environments, and thus present a limited performance on navigation problems in cluttered dynamic environments.
Overcoming this limitation of RL, meta-RL algorithms aim to learn control policies offline that can be quickly adapted to unseen environments or tasks.
Most of them are on-policy methods requiring massive data sets obtained from a large number of tasks~\cite{finn2017model, mishra2018simple, rothfuss2018promp}. 
To improve sample efficiency, 
off-policy methods have recently been proposed~\cite{rakelly2019efficient, zintgraf2020varibad, imagawa2022off}. 
A notable method is the probabilistic embeddings for actor-critic RL (PEARL), which disentangles task inference and control via a latent context variable to embed task information~\cite{rakelly2019efficient}.
 However, the meta-learned policy may be conservative in some situations, as information about various tasks is inevitably mixed during meta-training.

MPC is another sequential decision-making tool that inherently has an adaptation capability. 
MPC methods optimize control actions online in a receding horizon fashion. 
Once the environment or task changes, a new MPC problem can be solved to generate a control action that handles the new situation.  
In particular, MPC can be used in conjunction with various machine learning techniques to infer unknown parts of the system or environment model~\cite{hewing2020learning, koller2018learning, kabzan2019learning, aswani2013provably}. 
Unfortunately, learning-based MPC techniques are computationally demanding, particularly when handling complex problems.
Moreover, MPC optimizes an action sequence within a fixed short horizon, causing suboptimal, myopic behaviors in general.

Recently, there have been a few attempts to use MPC in model-based meta-RL or meta-learning methods.
Specifically, MPC has been used to exploit learned model information 
in real-time decision making~\cite{nagabandi2018learning, saemundsson2018meta, kaushik2020fast}.
However, none of them utilizes the actions generated by MPC in the training stage.
 Motivated by this observation, we propose a systematic approach to infusing MPC into the meta-RL training process for mobile robots in environments cluttered with moving obstacles. 
Our method is built upon PEARL and exploits its off-policy nature to generate transition samples using MPC when the robot detects certain events, such as risky movements near the obstacles, that can be effectively handled by MPC with its explicit use of robot dynamics. 
More precisely, the MPC controller is activated with probability $\varepsilon$ when predefined events occur.
This event-based activation of the MPC controller reduces the overall computational demand.
Unlike existing methods, the MPC module of our algorithm is carefully designed so that it learns the future motion of dynamic obstacles via Gaussian process regression (GPR)~\cite{rasmussen2003gaussian} and limits the risk of collision via conditional value-at-risk (CVaR) constraints~\cite{Rockafellar2002a}. Nevertheless, the dynamic nature of the environment as well as the limited horizon of MPC may lead to a suboptimal, short-sighted behavior. The proposed meta-learning strategy compensates for such limitations.

The meta-learned policy is then used to control the robot
without activating the MPC module to save computation time  
during meta-testing.
Thus, our controller is executed in a model-free manner.
We further propose an online adaptation scheme that frequently updates the latent context variable within the same episode. 
This enables the robot to adapt to a new task in a single trajectory, and thus it is useful in practical navigation problems in which it is difficult to repeat the same task.

Overall, our method benefits from the following complementary roles of MPC and PEARL:
\begin{itemize}
\item  The MPC module exploits the system dynamics of the robot when generating control actions in transition samples. 
It encourages exploration of potentially high-reward regions, thereby improving  learning stability and reducing the conservativeness of PEARL.

\item PEARL makes up for suboptimal MPC actions caused by obstacles not encountered within the limited prediction horizon.
Furthermore, it allows us to deactivate the MPC module during meta-testing, thereby significantly reducing computational effort in real-time motion control.
\end{itemize}

The performance of our method has been tested on two service robot scenarios with ($i$) synthetic  movements of obstacles in a restaurant and ($ii$) real-world pedestrian motion data on a sidewalk, using a variety of performance metrics to evaluate the safety and efficiency of trajectories. 
The results indicate that our method  outperforms all baselines in terms of learning efficiency and navigation quality.
Specifically, the travel time is reduced by more than $35\%$ compared to the baselines,  and the collision rate of the resulting trajectory is as low as $1\%$.

\section{Preliminaries}\label{sec:pre}

\subsection{Motion Control in Dynamic Environments}\label{sec:motion}

We consider a mobile robot navigating in a dynamic environment, such as a service robot for restaurants and airports.
The motion of the robot is assumed to be modeled by the following discrete-time nonlinear system:
\begin{equation}\label{robot_dyn}
x_{t+1} = f(x_t, u_t),
\end{equation}
where $x_t\in\mathcal{X}\subseteq\mathbb{R}^{n_x}$ and $u_t\in\mathcal{U}\subseteq\mathbb{R}^{n_u}$ are the robot's state and control input at stage $t$.

The robot aims to safely navigate
from an initial position
 to the desired goal position in an environment cluttered with obstacles.
Let $x_{t,i}^{d}\in\mathbb{R}^{n_{d}}$, $i=1, \dots, N_d$, and $x_{i}^{s}\in\mathbb{R}^{n_{s}}$, $i=1, \dots, N_s$, denote the state of dynamic  and static obstacles, respectively. 
The state of static obstacles is fixed in a single task, while the state of moving obstacles evolves according to unknown (stochastic) dynamics. As a common practice, we suppose that the obstacles' motion can be tracked with high accuracy. The fully observability assumption is valid in many real-world problems with a sufficient number of sensors. The motion control problem can then be formulated as a Markov decision process (MDP), defined as a tuple $\mathcal{T}:=(\mathcal{S}, \mathcal{U}, P,  c, \gamma)$,
where  $\mathcal{S} \subseteq \mathbb{R}^{n_x + n_d N_d + n_s N_s}$ and $\mathcal{U}$ are the state and action spaces, respectively, 
$P$ is the transition model, $c$ is a stage-wise cost function  of interest, and $\gamma \in (0, 1)$ is a discount factor.
The MDP state is defined as $s_t:=(x_t, x_{t,1}^d, \ldots, x_{t,N_d}^d, x_{1}^s,\ldots, x_{N_s}^s)$, concatenating the robot state and the states of the static and dynamic obstacles.
Given state $s_t$ and action $u_t$, $P(s_{t+1} | s_t, u_t)$ is the probability of being in $s_{t+1}$ at stage $t+1$.
Thus, $P$ encodes the mobile robot dynamics as well as the stochastic motion of dynamic obstacles.
Let $\pi$ denote the randomized Markov policy specifying a probabilistic rule of generating an action $u_t$ given the current state $s_t$. 
An optimal policy $\pi^\star(\cdot | \bm{s})$ can be obtained by solving the following MDP problem: 
\[
\min_{\pi} \mathbb{E}^\pi \left[
\sum_{t=0}^{\infty} \gamma^t c(s_t,u_t)~\middle|~s_0=\bm{s}
\right].
\] 
The stage-cost function $c$ is chosen as follows:
\begin{equation}\label{cost}
\begin{split}
&c(s_t, u_t) := \ell (x_t, u_t) \\
&+ w_d \bold{1}_{\{x_t \notin \mathcal{X}_t^d\}} + w_s \bold{1}_{\{x_t \notin \mathcal{X}^s\}} - w_g  \bold{1}_{\{x_t \in \mathcal{X}^g\}},
\end{split}
\end{equation}
where the loss function $\ell$ measures the control performance, 
$\mathcal{X}^d_t$ and $\mathcal{X}^s$ denote the safe regions with respect to dynamic and static obstacles to be defined, and
$\mathbf{1}_A$  returns $1$ if $A$ is true and $0$ otherwise.
Thus, the cost function penalizes the unsafe behaviors of the robot. 
Increasing the penalty parameters $w_d$ and $w_s$ encourages the robot 
to avoid collisions.  

\begin{figure}[t!]
    \centering
    \includegraphics[width=0.7\linewidth]{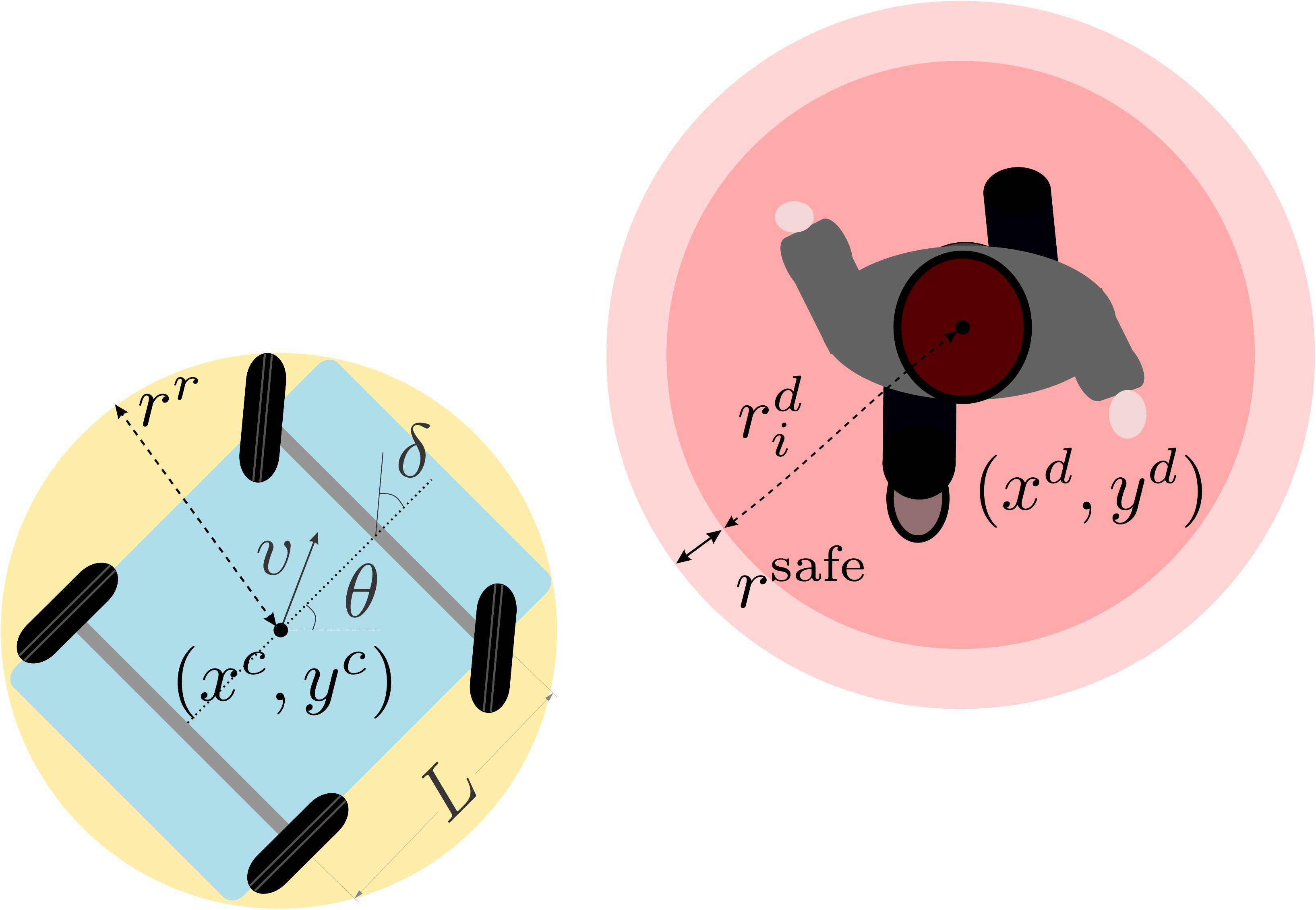}
    \caption{Illustration of parameters in the definition of $\epsilon^d$. Here, the human agent is considered as a dynamic obstacle.}
    \label{fig:safety}
        \vspace{-0.1in}
\end{figure}

The safe region with respect to the  dynamic  obstacles can be defined as
$\mathcal{X}^d_t:=\{x\in\mathbb{R}^{n_x} \mid h_d(x,x^{d}_{t,i})\geq \epsilon_i^d, \; i=1,\dots, N_{d}\}$.
Here, the safety function $h_d$ is chosen so that there is no collision if the function value is no less than $\epsilon_i^d$.  
In this work we set
$h_d (x, x_i^d) := \| p^r - p_i^d\|_2$,
 where $p^r$ and $p_i^d$ represent the center of gravities (CoGs) of
the robot and  dynamic obstacle $i$, respectively.
Accordingly, the threshold can be set to $\epsilon_i^d =r^r + r^d_i + r^{\mathrm{safe}}$, where $r^r$ and $r^d_i$ are the radii of balls covering the robot and the obstacle, respectively, and 
$r^{\mathrm{safe}}$ is a safety margin (Fig.~\ref{fig:safety}).
The safe region $\mathcal{X}^s$ with respect to static obstacles can be designed in a similar manner. 
Lastly, we let $\mathcal{X}^g:=\{x\in\mathbb{R}^{n_x} \mid \| p^r - p_{\mathrm{goal}}\|_2 \leq \epsilon^g\}$. Thus, the last term in \eqref{cost}
incentivizes the robot to reach a neighborhood of the goal point, chosen as the ball of radius $\epsilon^g$ centered at the goal.

\subsection{Off-Policy Meta-Reinforcement Learning}

A critical limitation of the optimal policy $\pi^\star$ is that it may no longer be effective when
the motion pattern of dynamic obstacles or
 the configuration of static obstacles changes. 
Thus, it is desired for a mobile robot to have the capability of quickly adapting to a new environment, possibly using past experiences. 
This motivates us to consider the motion control problem through the lens of meta-RL.

In meta-RL, a single MDP $\mathcal{T} := (\mathcal{S}, \mathcal{U}, P, c, \gamma )$ is considered as a \emph{task}.
Let $\{ \mathcal{T}^i\}_{i=1}^{N_{\mathrm{task}}}$ denote a collection of tasks, 
sampled from a given distribution of tasks
$p(\mathcal{T})$. 
In our navigation problem, 
 the motion pattern of dynamic obstacles and the configuration of static obstacles vary over different tasks. 
 Thus, the transition probability $P$ and the cost function $c$ are task-specific.

To enhance sample efficiency in meta-RL, it is critical to effectively use off-policy data.
As mentioned in Section~\ref{sec:intro},
PEARL~\cite{rakelly2019efficient}
is a state-of-the-art off-policy meta-RL algorithm.
PEARL exploits a \emph{latent  context variable} $\bold{z}$ 
encoding task information inferred through the experiences obtained from previous tasks.
Specifically, PEARL models the distribution over MDP with the latent context variable $\bold{z}$ and trains an encoder independently from the current policy to compress task information into the latent space and reuse off-policy data during meta-training.
Given a new task $\mathcal{T}^i$, $\bold{z}^i$ is sampled from the Gaussian prior $\mathcal{N}(0, I)$ to obtain transition data $(s_t,u_t,c_t,s_{t+1})$ executing a $\bold{z}^i$-conditioned policy $\pi_\theta (\cdot | s_t, \bold{z}^i)$. 
A collection of transition sample data $\bold{c}^i := \{\bold{c}_n^i\}_{n=1}^N:=  \{(s^{(n)}, u^{(n)}, c^{(n)}, s'^{(n)})\}_{n=1}^N$ is called the \emph{context}.
The context is then fed into a context encoder $q_\phi$, which maps $\bold{c}^i$ to the mean and variance of a Gaussian distribution.
 Specifically, the posterior of $\bold{z}$ given $\bold{c}_n^i$  is inferred as $\Psi_\phi(\bold{z} | \bold{c}_n^i) = \mathcal{N}(f^\mu_\phi(\bold{z} | \bold{c}_n^i), f^\sigma_\phi(\bold{z} | \bold{c}_n^i))$. 
 Motivated by the observation that each MDP can be inferred from the set of transitions,
 PEARL performs a permutation-invariant posterior update for $\bold{z}$ given context $\bold{c}^i$ as follows:
    $q_\phi(\bold{z}|\bold{c}^i) \propto \prod_{n=1}^N \Psi_\phi(\bold{z}| \bold{c}^i_n )$.
With new $\bold{z}^i$ sampled from the posterior, 
the aforementioned process is repeated to obtain a better inference of the latent context.

\begin{figure}[t!]
    \centering
    \includegraphics[width=0.95\linewidth]{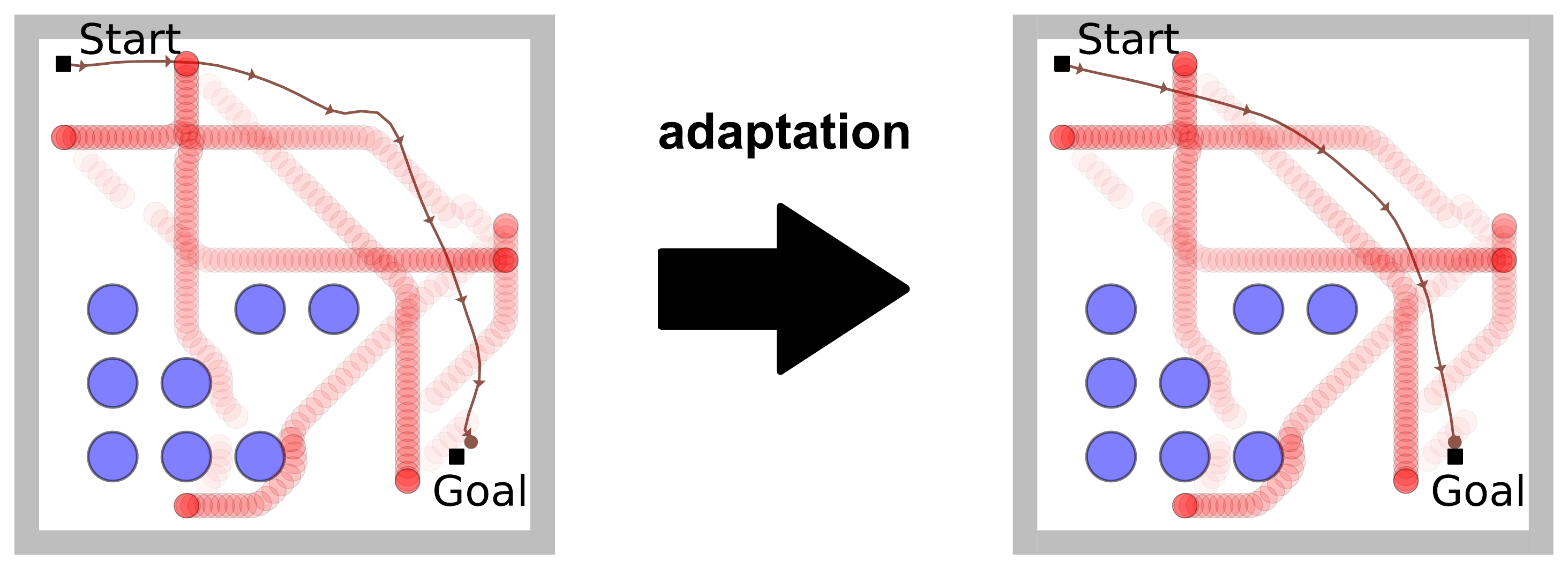}
\caption{The trajectory generated by PEARL (left) before and (right) after adaptation over 2 episodes. 
The movements of dynamic obstacles and
the configuration of static obstacles  are shown in red and blue, respectively.
}
\label{fig:pearl_traj}
    \vspace{-0.1in}
\end{figure}

{A meta-RL example of a car-like robot  is shown in Fig.~\ref{fig:pearl_traj}.\footnote{Details about the experiment can be found in Section~\ref{sec:result}.}}
Even after adaptation, the PEARL policy is unable to drive the robot to take a near-optimal path that passes through the central region of the environment. 
 This example indicates that PEARL policies may induce overly conservative behaviors to bypass regions cluttered with moving obstacles. 

\begin{figure}[t]
    \centering
    \includegraphics[width=0.95\linewidth]{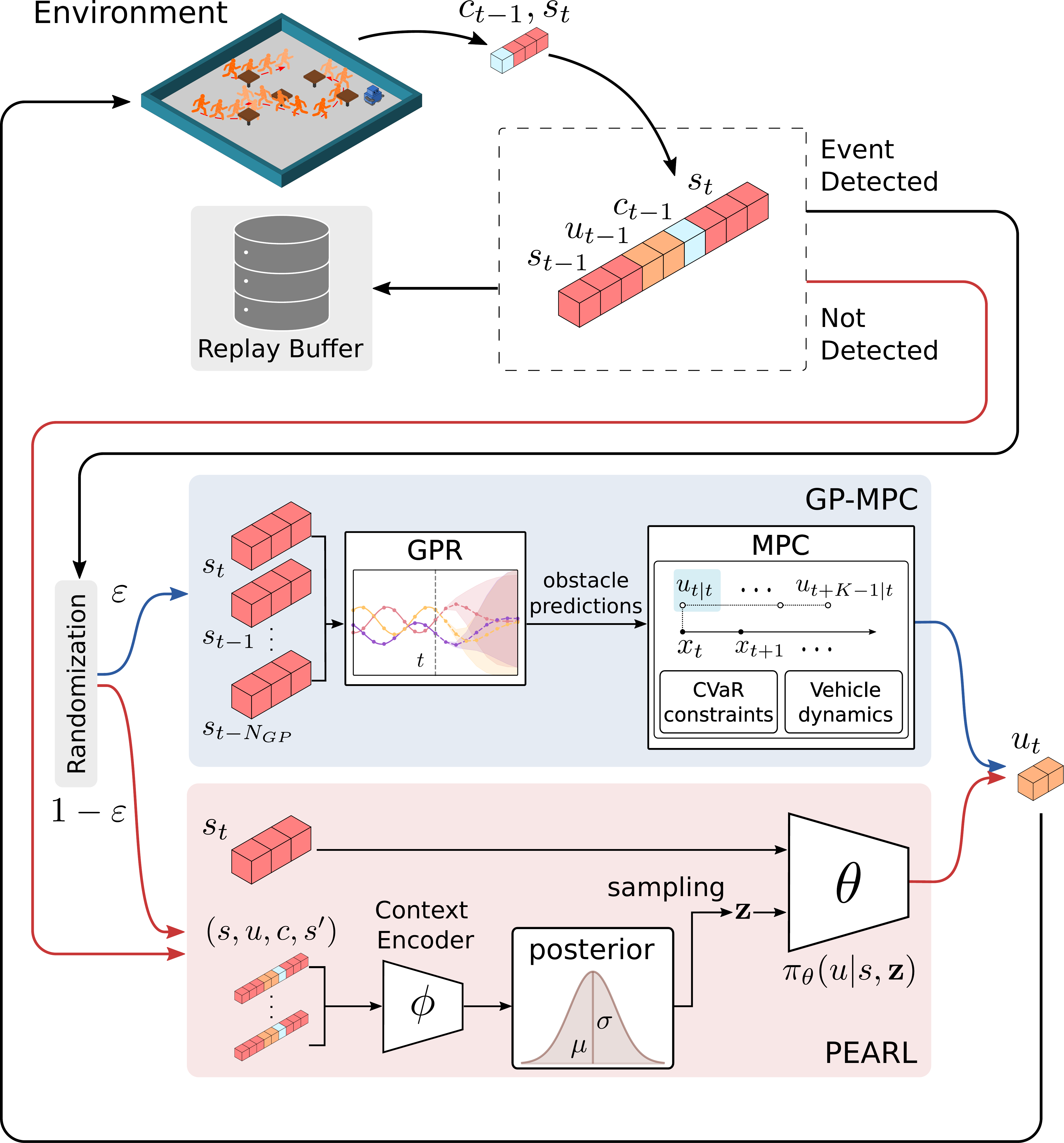}
    \caption{Overview of MPC-PEARL. }
    \label{fig:overview}
    \vspace{-0.1in}
\end{figure}

\section{Infusing MPC into Meta-RL}
\label{sec:infuse}

The conservativeness of PEARL in a dynamic environment impedes the exploration of high reward regions. To resolve the limited adaptation capability and navigation performance of PEARL, we propose combining it with learning-based MPC.
Fig.~\ref{fig:overview} shows an overview of our method, which we call MPC-PEARL. 
Our method exploits the off-policy nature of PEARL to generate transition samples using MPC when a predefined event occurs.\footnote{In our motion control problem, we consider the following two events that are ineffectively handled by PEARL: (Event 1) collision with a dynamic obstacle, and (Event 2) reaching the neighboring region of the goal point.}
In each time step, the robot observes the current state $s$ and probabilistically decides whether to use MPC when the event is detected.
If the MPC module is activated, a carefully designed MPC problem is solved using the GPR result of inferring the motion of obstacles.
Otherwise, the original PEARL algorithm is used to execute the policy $\pi_\theta(\cdot | s, \bold{z})$ conditioned on the latent context variable $\bold{z}$.
Note that the randomization layer induces MPC-PEARL to learn frequently from actions generated by MPC.

\subsection{MPC-PEARL Algorithm}\label{sec:metarllearning}

\begin{algorithm}[t]
\DontPrintSemicolon
\caption{MPC-PEARL meta-training}
\label{alg:PEARL_MPC_train}
\SetKw{to}{to}
\SetKw{and}{and}
\SetKw{Input}{Input:}

\Input Set of training tasks $\{\mathcal{T}^i \}_{i = 1}^{N_{\mathrm{task}}}\sim p(\cdot)$\;
Initialize replay buffers $\mathcal{H}^{(i)}$, and network weights $\phi, \theta, \psi$ \; 
\For{iteration $1,2,\ldots$}{
    \ForEach(\tcp*[h]{Data Collection}){$\mathcal{T}^i$}{
        Initialize $\bold{c}^i = \{ \}$ and sample $\bold{z}^i \sim \mathcal{N}(0, I)$ \;
        \For{episode $=1$ \to $E$}{
            \For{$t=0$ \to $T - 1$}{
                \eIf{Event occurs and $\mathrm{rand} \leq \varepsilon$}{
                Build GP dataset $\mathcal{D}_t =  \{({x^d_{t-l}, x^d_{t-l+1}-x^d_{t-l}})\} _{l=1}^{N_{\mathrm{GP}}}$\;
                Predict $\{ \hat{x}^d_{t+k} \}_{k=1}^K$ via GPR\;
                Obtain $u_t$ by solving \eqref{LBMPC}
                }{
                $u_t \sim \pi_{\theta}(\cdot | s_t, \bold{z}^i)$\;
                }
                Apply $u_t$ and observe $(s_t, u_t, c_t, s_{t+1})$\;
                Add $(s_t, u_t, c_t, s_{t+1})$ to $\mathcal{H}^{i}$\;
            }  
         Update context $\bold{c}^i = \{(s^{(n)}, u^{(n)}, c^{(n)}, s'^{(n)})\}_{n=1}^{N} \sim \mathcal{H}^{i}$\;
         Sample $\bold{z}^i \sim q_\phi(\cdot | \bold{c}^i)$ \;
        }
        
    }
    \For(\tcp*[h]{Training}){$\ell=1$ \to $N_{\mathrm{train}}$}{
    	\ForEach{$\mathcal{T}^i$}{
            Sample context $\bold{c}^i \sim {\mathcal{S}}_c(\mathcal{H}^{i})$ and batch $\mathcal{B}^i \sim \mathcal{H}^{i}$\;
            Sample $\bold{z}^i \sim q_\phi(\cdot | \bold{c}^i)$\;
                    }
        $\phi \gets \phi - \alpha_\phi \nabla_\phi \sum_i \mathcal{L}_\phi^i$, $\theta \gets \theta - \alpha_\theta \nabla_\theta \sum_i \mathcal{L}_\theta^i$\;
        $\psi \gets \psi - \alpha_\psi \nabla_\psi \sum_i \mathcal{L}_\psi^i$\;
    }
}
\end{algorithm}

The overall meta-training process of MPC-PEARL is described in Algorithm~\ref{alg:PEARL_MPC_train}. 
A set of training task $\{\mathcal{T}^i\}_{i=1}^{N_\mathrm{task}}$ sampled from the task distribution $p(\mathcal{T})$ is taken as the input. 
Each iteration of the algorithm consists of the data collection stage (line 4--17) and the training stage (line 18--{23}).

In the data collection stage,
for each task $\mathcal{T}^i$, transition samples are collected for a fixed number of episodes to build a task-specific replay buffer $\mathcal{H}^i$ using the policy $\pi_\theta$ and the context encoder $q_\phi$ trained in the previous iteration.
The original PEARL algorithm uses $\pi_\theta (\cdot | s_t, \bold{z}^i)$, conditioned on the current latent variable $\bold{z}^i$, 
to generate a control action when collecting transition samples (line 13). 
However, another policy can be used instead as PEARL is an off-policy algorithm. 
Exploiting this feature, 
we propose to use a randomization layer that activates MPC with  probability $\varepsilon$ to generate an  action 
if a predefined event occurs.
In particular, our MPC module is carefully designed to predict the motion of dynamic obstacles via GPR and to avoid collisions via CVaR constraints as presented in Section~\ref{sec:LMPC} {(line 9--11)}.
Note that the MPC complements PEARL, which is a model-free algorithm, by producing a high-performance control action through effective use of the system model~\eqref{robot_dyn} in the current and predicted situations.
At the end of each episode, the context $\bold{c}^i$ is sampled from the replay buffer $\mathcal{H}^i$ (line 16), and accordingly, a new latent context variable $\bold{z}^i$ is sampled from the posterior $q_\phi(\cdot| \bold{c}^i)$ (line 17). 
This process is repeated to construct replay buffers for all tasks.

In the training stage, 
the policy $\pi_\theta$ and the context encoder $q_\phi$ are trained using the replay buffers.
For each task $\mathcal{T}^i$, a context $\bold{c}^i$ is drawn from the replay buffer $\mathcal{H}^i$ using a sampler $\mathcal{S}_c$,\footnote{Similar to the original PEARL implementation, $\mathcal{S}_c$ is chosen to generate transition samples uniformly from the most recent data collection stage.}
and an RL minibatch  $\mathcal{B}^i = \{ (s^{(m)}, u^{(m)}, c^{(m)}, s'^{(m)})\}_{m=1}^M$ is sampled independently. 
Using separate data for training the encoder and the policy yields to perform all updates for a fixed latent context $\bold{z}^i$, sampled from $q_\phi (\cdot | \bold{c}^i)$ in line 21, which can be considered as an additional state. 
The policy is trained using soft actor-critic (SAC)~\cite{haarnoja2018soft}, where the actor loss $\mathcal{L}_{\theta}^{i}$ and the critic loss $\mathcal{L}_{\psi}^{i}$ are defined as
$\mathcal{L}_{\theta}^{i} = \frac{1}{M}\sum_{m=1}^{M}  D_{\mathrm{KL}} \left( \pi_\theta (u|s^{(m)},\bold{z}^i)  \left\|   \frac{\exp(Q_\psi(s^{(m)},u,\bold{z}^i))}{\mathcal{Z}_\psi(s^{(m)})}\right.  \right)$ and
$\mathcal{L}_{\psi}^{i}$ $= \frac{1}{M}\sum_{m=1}^{M} \left [c^{(m)} + \gamma\bar{V}({s '}^{(m)}, \bold{z}^i) - Q_\psi(s^{(m)}, u^{(m)}, \bold{z}^i) \right]^2$, respectively.
Here, $\bar V$ is the target value function.
The actor loss is designed to minimize the KL divergence between the current policy and the greedy soft policy, while the critic loss aims to reduce the temporal difference error for the soft Q-function.
Furthermore, the context encoder network weight $\phi$ is trained by minimizing
$\mathcal{L}_{\phi}^{i} = \mathcal{L}_{\psi}^{i} + \beta D_{\mathrm{KL}}\left( q_\phi(\bold{z} | \bold{c}^i)  \left\| \mathcal{N}( 0, I)\right.  \right)$,
which is formulated using the amortized variational inference approach~\cite{rezende2014stochastic, alemi2016deep} with the standard Gaussian prior $\mathcal{N}(0, I)$. 
As in PEARL, the critic loss is additionally considered in the encoder loss since small $\mathcal{L}_\psi^i$ implies that  $Q_\psi$ is close to the optimal one, and thus the encoder successfully identifies the given task.
Finally, the parameters can be updated by stochastic gradient descent or an adaptive optimization method such as Adam~\cite{kingma2015iclr}.

\begin{algorithm}[t]
\DontPrintSemicolon
\caption{MPC-PEARL meta-testing}
\label{alg:PEARL_MPC_test}
\SetKw{Input}{Input:}
\SetKw{to}{to}
\SetKw{and}{and}
\Input Test task $\mathcal{T}\sim p(\cdot)$\;
Initialize $\bold{c}^\mathcal{T} = \{ \}$ \;
        \For{episode $=1, 2, \ldots$}{
            \For{$t=0$ \to $T - 1$}{
                \If{$t \; \mathrm{mod} \; t_{\mathrm{update}} \equiv 0$}{
            Sample $\bold{z}^\mathcal{T} \sim q_\phi (\cdot |  \bold{c}^\mathcal{T})$ 
        }
                Sample $u_t \sim \pi_{\theta}(\cdot | s_t, \bold{z}^\mathcal{T})$\;
                Apply $u_t$ and observe $(s_t, u_t, c_t, s_{t+1})$\;
        $\bold{c}^\mathcal{T} \leftarrow \bold{c}^\mathcal{T} \cup \{(s_t, u_t, c_t, s_{t+1})\}$ \;
        }
}
\end{algorithm}

The meta-testing process for a task $\mathcal{T}$ sampled from the task distribution is presented in Algorithm~\ref{alg:PEARL_MPC_test}.
As opposed to meta-training, the actions are generated from the learned policy without explicitly executing MPC. 
As meta-training benefits from using MPC with randomization, the learned policy is likely to execute actions that are no worse than those generated by MPC.
Moreover, by omitting the MPC module during meta-testing, 
a significant amount of computation time can be saved, 
as only a simple function evaluation is performed online instead of costly  numerical optimization. Unlike meta-training, the posterior of the latent context variable $\bold{z}^\mathcal{T}$ is updated every $t_{\mathrm{update}}$ step using the transition data collected so far. A new context variable is sampled from the encoder $q_\phi$ to infer the current task using the acquired real-time information (lines 5 and 6).
This online adaptation scheme, performed within a single trajectory, is crucial for some practical applications where each episode is less likely to be repeated due to consistently evolving environments.

\subsection{Learning-Based MPC with CVaR Constraints}\label{sec:LMPC}

As described in Section~\ref{sec:metarllearning}, PEARL suffers from 
conservativeness and inefficiency in the navigation environment.
Furthermore, PEARL does not use the known system dynamics that is often available for various practical robots.
To make up for the limitation of PEARL, 
we propose to adopt a learning-based MPC technique that solves the given task and uses model information in a receding horizon fashion:\footnote{Here, $x_{\tau | t}$ denotes the system state at stage $\tau$ predicted at stage $t$. Thus, it may be different from $x_\tau$, the actual state at stage $\tau$.}
\begin{subequations}
\begin{align}
\min_{\mathbf{u}} \;& J(x_t,\mathbf{u}) = \ell_f(x_{t+K|t}) + \sum_{k=0}^{K-1}\ell(x_{t+k|t},u_{t+k|t})\\
\mbox{s.t. }& x_{t+k+1|t} = f(x_{t+k|t},u_{t+k|t})\label{mpc_dyn}\\
&x_{t|t} = x_t\label{const_init}\\
& x_{t+k|t}\in \mathcal{X}^d_{t+k} \cap \mathcal{X}^s   \label{const_xd}\\
&u_{t+k|t}\in \mathcal{U}, \label{const_u}
\end{align}\label{MPC}%
\end{subequations}
where $\mathbf{u}=(u_{t|t}, \dots,u_{t+K-1|t})$ is a control input sequence over the $K$ prediction horizon, constraints~\eqref{mpc_dyn} and~\eqref{const_u} must be satisfied for $k=0,\dots,K-1$ and \eqref{const_xd} must hold for $k=0,\dots,K$. 
Here, the terminal cost function $\ell_f$ takes into account the costs occurring beyond the prediction horizon.
The state constraint~\eqref{const_xd} ensures that the robot is located in the safe region.
The MPC problem at stage $t$ is solved in a receding horizon fashion starting from $x(t)$ followed by the execution of the first entry of the optimal input sequence $\bold{u}^\star$, i.e., $u_t = {u}^\star_{t|t}$.
Unfortunately, the MPC problem in the above form is impossible to solve because 
in general the safe region $\mathcal{X}_{t+k}^d$ is unknown; it includes information about the future motion of dynamic obstacles. 

\subsubsection{Learning the motion of obstacles via GPR}

To infer the future state $x^d_{t+k,i}$ of dynamic obstacle $i$, we use GPR~\cite{rasmussen2003gaussian}, 
one of the most popular non-parametric methods for learning a probability distribution over all possible values of a function. 
 For simplicity, we suppress the subscript $i$. 
 GPR is performed online using a training dataset $\mathcal{D}_t = \{x_{t-l}^d, v_{t-l}^d\}_{l=1}^{N_{\mathrm{GP}}}$ constructed from $N_{\mathrm{GP}}$ most recent observations of the obstacles' state transitions, where the input is the state at time $t-l$ and the output is the difference between two consecutive states, i.e. $v_{t-l}^d :=x_{t-l+1}^d - x_{t-l}^d$.
The GP approximation of $x^d_{t+k}$ performed at stage $t$ is obtained as
$\hat{x}^d_{t+k} \sim\mathcal{N}(\mu_{t+k}, \Sigma_{t+k})$. Here, the mean $\mu_{t+k}\in\mathbb{R}^{n_d}$ and the covariance $\Sigma_{t+k}\in\mathbb{R}^{n_d \times n_d}$ are computed by propagating the state vector using current observation $x^d_{t}$ and GP approximation of $v^d_{t+k}$. The detailed update rule for the mean and covariance can be found in Appendix. A potential issue arising from using GPR is its computational complexity. However, this issue can be easily avoided by employing a separate GP model for each dynamic obstacle and learning the obstacles in parallel.

\subsubsection{CVaR constraints for safety}
Using the inferred state, the safe region can be predicted as 
$\hat{\mathcal{X}}^d_{t+k} :=\{x\in\mathbb{R}^{n_x} \mid h_d(x, \hat{x}^{d}_{t+k,i})\geq \epsilon_i^d \; \forall i \}$.
However, due to the stochastic nature of $\hat{\mathcal{X}}^d_{t+k}$,
imposing the deterministic constraint $x_{t+k |t} \in \hat{\mathcal{X}}^d_{t+k}$ is likely to cause infeasibility or an overly conservative solution. 
As a remedy, a probabilistic constraint can be used instead. 
In particular, we use a CVaR constraint.
The CVaR of a random loss $X$ is its expected value within the $(1-\alpha)$ worst-case quantile and is defined as $\mathrm{CVaR}_\alpha[X]:=\min_{\xi\in\mathbb{R}} \mathbb{E}[ \xi+(X-\xi)^+ / (1-\alpha)]$, where $\alpha\in(0,1)$ and $(x)^+ = \max\{x,0\}$~\cite{Rockafellar2002a}.
CVaR is a convex risk measure and discerns a rare but catastrophic event in the worst-case tail distribution unlike chance constraints.

In our problem, we let
$L(x_t, \hat{x}_t^d ) := \max_{i = 1, \ldots, N_d} \{ \epsilon_i^d - h_d (x_t, \hat{x}_{t,i}^d)  \}$
denote the loss of safety at stage $t$. 
Note that $L(x_{t+k |t}, \hat{x}_{t+k}^d ) \leq 0$ if and only if $x_{t+k |t}$ lies  in  the predicted safe region.
We replace the deterministic constraint $x_{t+k |t} \in \hat{\mathcal{X}}^d_{t+k}$ with the following CVaR constraint:
\begin{equation} \nonumber
\begin{split}
&\mathrm{CVaR}_\alpha [L(x_{t+k |t}, \hat{x}_{t+k}^d )]=\\
&\min_{\xi\in\mathbb{R}} \bigg ( \xi+\frac{1}{1-\alpha}\mathbb{E}\big[\big(L(x_{t+k |t},\hat{x}_{t+k}^d) - \xi\big)^+\big] \bigg )\leq \delta_{\mathrm{CVaR}},
\end{split}
\end{equation}
where $\delta_{\mathrm{CVaR}}$ is a user-specified non-negative threshold representing
the maximum tolerable risk level. When $\delta_\mathrm{CVaR}=0$, the robot is not exposed to any risk, which can be overly restrictive. Increasing $\delta_\mathrm{CVaR}$ relaxes the conservativeness of the safety constraint, allowing the robot to take some risk.

 Using a dataset $\{\hat{x}_{t+k}^{d, (m)} \}_{m=1}^{M_{\mathrm{SAA}}}$ sampled from the probability distribution of $\hat{x}_{t+k}^d$ inferred via GPR,   
 the CVaR constraint can be approximated by sample average approximation (SAA)
as in~\cite{hakobyan2019risk}, 
 and we arrive at the following GP-MPC formulation:
\begin{subequations}
\begin{align}
\min_{\mathbf{u},\mathbf{\xi}} \;\;&J(x_t,\mathbf{u})\\
\mbox{s.t.} \;\; & 
\xi_k+  \sum_{m=1}^{M_{\mathrm{SAA}}}\frac{\big( L(x_{t+k|t}, \hat{x}^{d, (m)}_{t+k}) - \xi_k\big)^+}{M_{\mathrm{SAA}}(1-\alpha)} \leq \delta_{\mathrm{CVaR}} \label{const_xd1}\\
&x_{t+k|t}\in \mathcal{X}^s  \label{const_x1}\\
& \mbox{\eqref{mpc_dyn}, \eqref{const_init}, and \eqref{const_u}.}
\end{align}\label{LBMPC}%
\end{subequations}
In general, the MPC problem~\eqref{LBMPC}
is a nonconvex optimization problem.
A locally optimal solution can be obtained 
using interior-point methods or sequential quadratic programming algorithms, among others~\cite{nocedal2006numerical}.
The SAA approach provides a provable asymptotic guarantee of satisfying the original safety risk constraint as shown in~\cite{hakobyan2019risk}. 
However, solving the MPC problem with a large sample size can be computationally demanding. 
In our experiments, the resulting controller with
a reasonable sample size $(M_{\saa} = 100)$ presents a satisfactory performance, demonstrating the validity of the SAA approach. 
The issue of limited sample sizes and learning errors can be systematically addressed by using distributionally robust optimization and control that ensure safety in the worst-case situations~\cite{Hakobyan2021}.


\begin{table}[t]
	\centering
	\caption{Hyperparameters for MPC-PEARL.}
	\begin{tabular}{l | *{3}{c}}
		\hline
		Hyperparameter  			  			&   \\
		\hline \hline
		optimizer      							& Adam~\cite{kingma2015iclr}         \\
		actor/critic/encoder learning rate  							& $3\times 10^{-4}$ \\ 
		discount factor $(\gamma)$       	& $0.99$\\
		latent space dimension                          & 10 \\
		replay buffer size       				& $10^6$           \\
		target smoothing coefficient $(\alpha)$	& $5 \times 10^{-3}$   \\
		temperature parameter			& 0.01\\
		\# of hidden layers 				& 3 (fully connected)         \\
		\# of hidden units per layer  		& 300	\\
		\# of encoder hidden units per layer & 200 \\
		\# of samples per minibatch 		& 256    \\
		\# of samples per encoder minibatch & 64 \\
		nonlinearity 							& ReLU  \\
		\hline
	\end{tabular}
	\begin{tabular}{ *{2}{c}}
	\hline
	\hline
\end{tabular}
	\label{tab:hyp}
	\vspace{-0.1in}
\end{table}

\begin{table}[t]
	\centering
	\caption{Hyperparameters for GrBAL.}
	\begin{tabular}{l | *{3}{c}}
		\hline
		Hyperparameter  			  			&   \\
		\hline \hline
		optimizer      							& Adam~\cite{kingma2015iclr}         \\
		learning rate  							& $10^{-3}$ \\
		inner learning rate                    & $10^{-2}$ \\
		discount factor $(\gamma)$       	& $0.99$\\
		MPC horizon length                & $10$ \\
		\# of samples per minibatch & 256 \\
		\# of hidden layers 				& 3 (fully connected)         \\
		\# of hidden units per layer  		& 256	\\
		\# of samples per adaptation 		& 4    \\
		\# of tasks per metabatch & 24 \\
		nonlinearity 							& ReLU  \\
		\hline
	\end{tabular}
	\begin{tabular}{ *{2}{c}}
	\hline
	\hline
\end{tabular}
	\label{tab:hyp_grbal}
	\vspace{-0.1in}
\end{table}

\section{Simulation Results}\label{sec:result}

In this section, we evaluate the performance of our method through the motion control of a service robot operating in two environments: ($i$) a simulated restaurant and ($ii$) a real-world sidewalk.
All of our experiments consider a car-like robot with a simple front wheel steering bicycle model~\cite{rajamani2011vehicle}, as illustrated in Fig.~\ref{fig:safety}. The state consists of 
the Cartesian coordinates of the robot's CoG, which is $(x_t^c, y_t^c)$, and the heading angle $\theta_t$.
The control input is chosen as the velocity $v_t$ and the steering angle of the front wheel $\delta_t$. 
The action space is set to $\mathcal{U}:=[0,v_{\mathrm{max}}]\times [-\frac{\pi}{2},\frac{\pi}{2}]$ with $v_{\mathrm{max}} = 7$ in the restaurant environment and $v_{\mathrm{max}} = 1.8$ in the sidewalk environment.
 Then, the equations of motion  are given by
\begin{align*}
x^c_{t+1} & = x^c_{t} + T_s v_t \cos \big(\theta_t + \beta_t\big)\\
y^c_{t+1} & = y^c_{t} + T_s v_t \sin \big(\theta_t + \beta_t\big)\\
\theta_{t+1} & = \theta_t + T_s \frac{v_t\cos (\beta_t)}{L} \tan (\delta_t),
\end{align*}
where $T_s$ is the sample time and $L = 0.5$ is the distances between the front and rear axles. The slip angle $\beta_t$ is defined as
$\beta_t := \tan^{-1}  (\frac{\tan \delta_t}{2})$. For the restaurant environment, $T_s = 0.05$ was used, while $T_s = 0.04$ was chosen for the sidewalk environment.
The ball covering the robot has a radius of $r^r = L/2$.
We also let $r^s_i = 0.5$ and $r^d_i = r^{\mathrm{safe}}= 0.25$. 
The safety thresholds were selected as $\epsilon^d_i = r^r + r_i^d + r^{\mathrm{safe}}$, $\epsilon^s_i = r^r + r_i^s + r^{\mathrm{safe}}$, and $\epsilon^g = 2r^r$.

 For both environments, the MPC module is activated with a probability of $\varepsilon\in\{0.2, 0.5, 0.8, 1\}$ when the events occur.
The cost parameters were chosen  to speed up navigation with limited control energy ($Q= \mathrm{diag}[1,1,0]$, $R=0.2 I_{n_u}$), to incentivize the robot to move close to the target ({$w_g = 10$}) and give a large collision penalty  ({$w_s = w_d = 20$}). 
In the MPC problem, the terminal cost function was chosen as $\ell_f(x):=\|x - x_{\mathrm{goal}}\|^2_{Q_f}$ with $Q_f = 20 Q$.
Throughout all experiments the MPC problem is solved for a planning horizon of $K=10$ using a fast and reliable optimization tool called FORCES PRO~\cite{FORCESPro}. The GPR is performed using the latest $N_{\mathrm{GP}}=10$ observations for predicting the motion of the closest $N_{\mathrm{adapt}}=2$ obstacles. 
For safety, we consider the approximate CVaR risk constraint with $\alpha=0.95$ and risk threshold $\delta_\mathrm{CVaR}=0.01$ using $M_{\mathrm{SAA}}=100$ samples. 

\begin{figure}[t!]
\centering
\begin{subfigure}[b]{\linewidth}
\centering
\includegraphics[width=\linewidth]{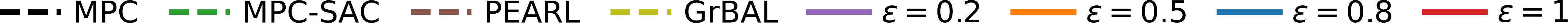}
\end{subfigure}\vspace{0.05in}\\
\begin{subfigure}[b]{0.32\linewidth}
    \centering
    \includegraphics[width=\linewidth]{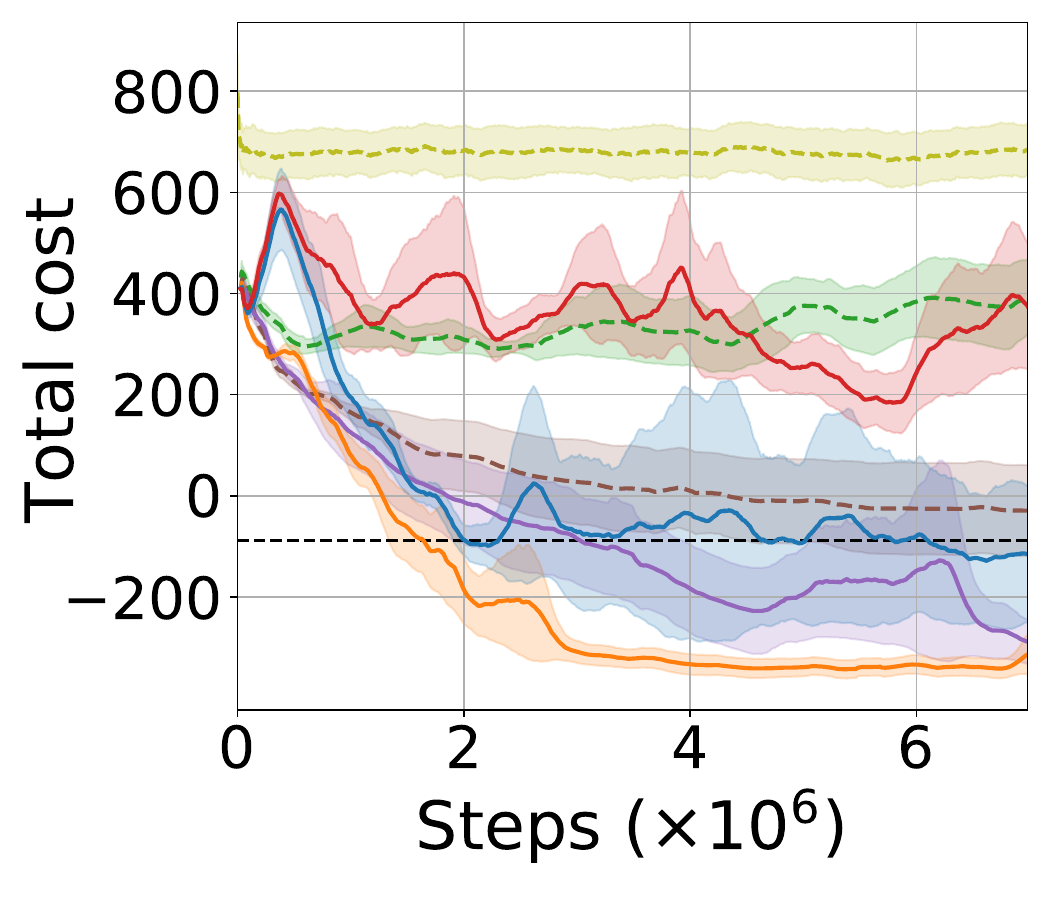}
    \caption{ }
    \label{fig:return}
\end{subfigure}
\begin{subfigure}[b]{0.32\linewidth}
    \centering
    \includegraphics[width=\linewidth]{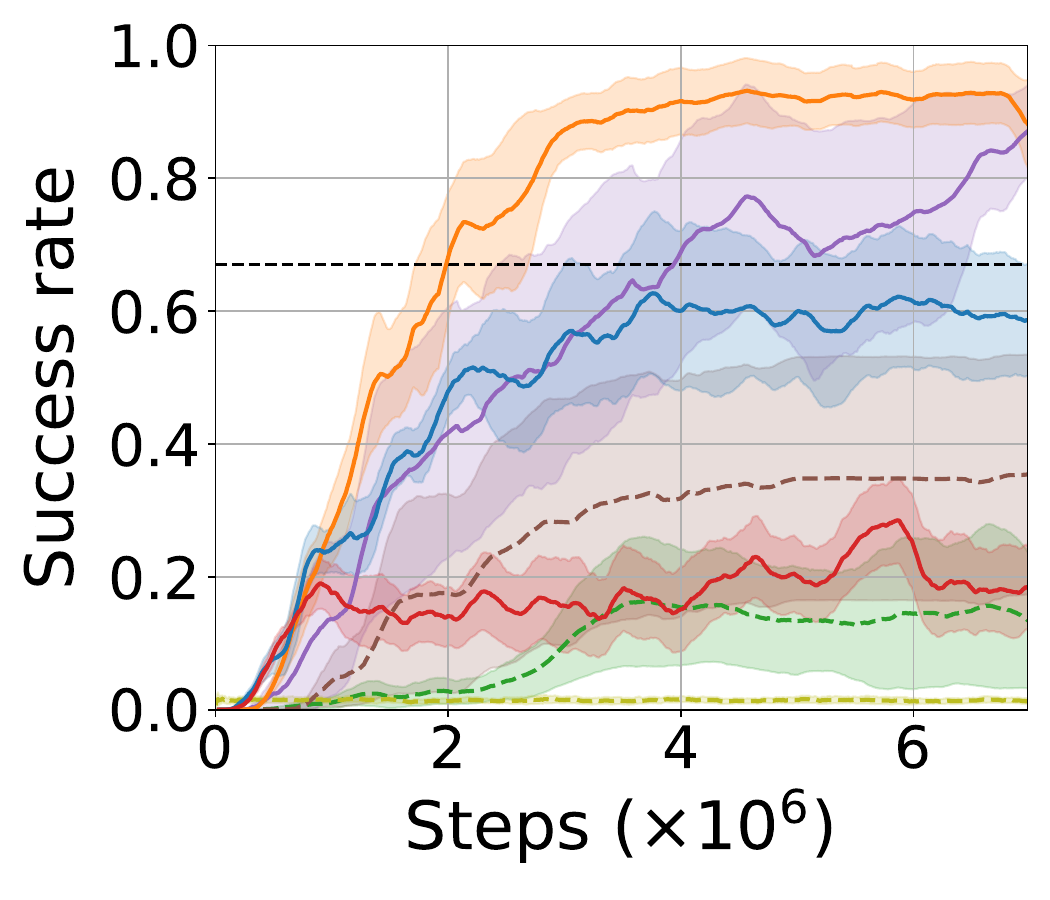}
    \caption{ }
    \label{fig:collision_free_success}
\end{subfigure}
\begin{subfigure}[b]{0.32\linewidth}
    \centering
    \includegraphics[width=\linewidth]{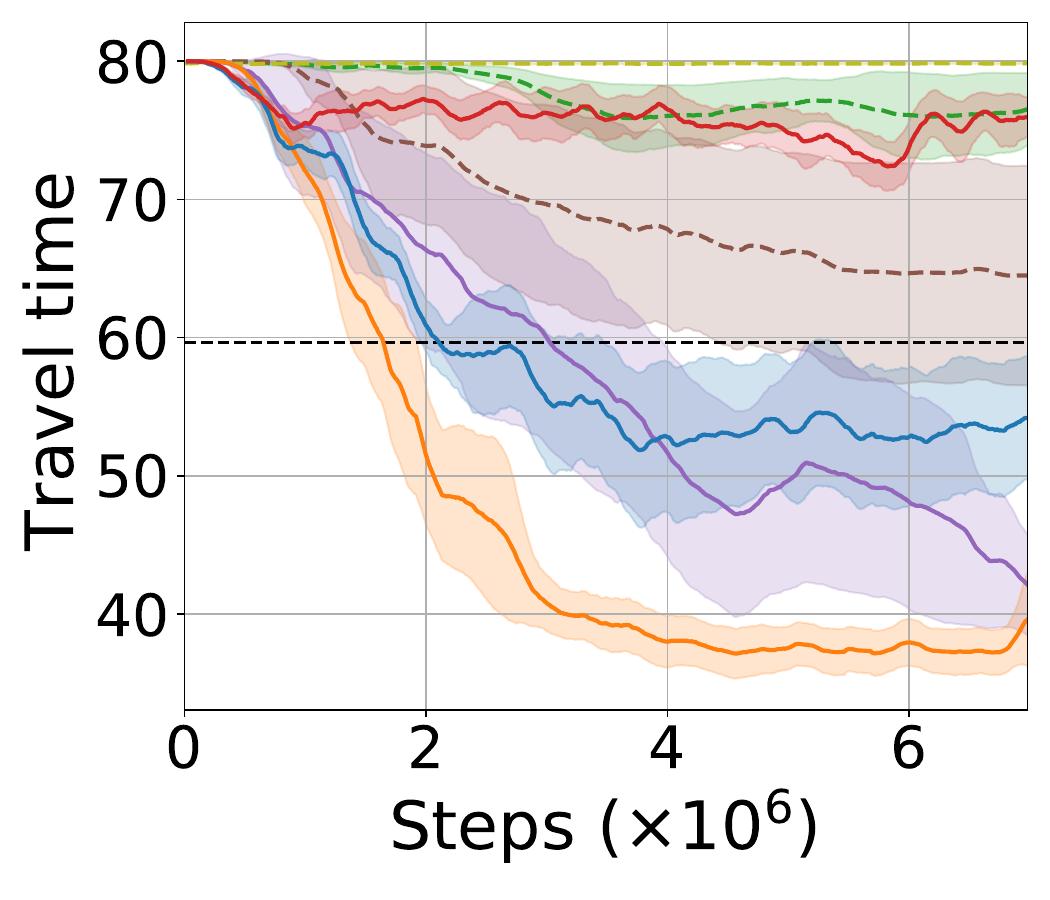}
    \caption{ }
    \label{fig:travel}
\end{subfigure}
\caption{Comparison of various methods in terms of
(a) the total cost, (b) success rate, and (c) travel time, evaluated after adaptation over two trajectories. 
The solid lines are the results of MPC-PEARL with different $\varepsilon$'s.
The lines indicate the mean of the corresponding evaluation metric across 7 different seeds, while the shaded region represents half a standard deviation.
}
\label{fig:results}
\vspace{-0.1in}
\end{figure}

We compare our method to PEARL, GP-MPC, MPC-SAC and GrBAL~\cite{nagabandi2018learning}, using the following metrics:\footnote{MPC-SAC uses the same infusion scheme as that in MPC-PEARL.  The hyperparameters of GrBAL are summarized in Table~\ref{tab:hyp_grbal}.}
\begin{itemize}
\item \emph{Success Rate (SR)}: the proportion of tasks in which the robot reaches the goal without any collision.

\item \emph{Travel Time (TT)}: the time for the robot to reach the goal without any collision during navigation.

\item \emph{Collision-Free Rate (CFR)}: the proportion of tasks in which no collision occurs.

\item \emph{Success Weighted by Path Length (SPL)}: a comprehensive score that considers the accomplishment of the task and the quality of the trajectory, as defined in~\cite{anderson2018evaluation}.

\item \emph{Mean Goal Distance (MGD)}: the average Euclidean distance between the robot and the goal.
 
\end{itemize}

All the experiments were conducted on a PC with Intel Core i9 10940X @ 3.30 GHz and an NVIDIA GeForce RTX 3090. 
The hyper-parameters used in our MPC-PEARL and PEARL implementation are reported in Table~\ref{tab:hyp}.
The source code of our implementation is available online\footnote{https://github.com/CORE-SNU/MPC-PEARL.git}.

\subsection{Restaurant Environment}\label{sec:rest}

We first consider a restaurant environment with synthetic movements of  obstacles.
The restaurant has a square floor plan with circular static and dynamic obstacles to model the table and customers, respectively.
We designed 8 different table configurations and 15 different behaviors of dynamic obstacles.
Each of the combinations forms a task, then the total number of tasks is $N_{\mathrm{task}} = 120$.
MPC-PEARL and PEARL are trained on these tasks by updating the network parameters $N_{\mathrm{train}}=750$ times per iteration. 
The results are tested on $24$ unseen tasks by adapting via two episodes.

Fig.~\ref{fig:results} displays the learning curves for MPC-PEARL with various MPC activation probabilities $\varepsilon$. 
As shown in Fig.~\ref{fig:results} (a), our method outperforms PEARL in terms of converged cost, sample efficiency, and stability for $\varepsilon$ between $0.2$ and $0.8$. The cases with $\varepsilon = 0.2, 0.5$ show the best performance. While $\varepsilon =0.8$ exhibits the initial learning speed comparable with 
$\varepsilon = 0.2, 0.5$, its final performance is worse than the two. 
This is because, compared to $\varepsilon = 0.2, 0.5$,  
the case with $\varepsilon = 0.8$ has fewer opportunities to discover control actions that are better than MPC actions.
Similarly, the unstable training performance of $\varepsilon = 1$ is not surprising since action diversity is limited when an event occurs in this setting.
This confirms the need for a randomization layer in our method.
Fig.~\ref{fig:results} (b) and (c) show SR and TT, respectively.
Our method with $\varepsilon = 0.2$ outperforms GP-MPC in terms of SR and TT. GrBAL presents the worst performance in terms of all metrics.
As GrBAL is a pure model-based method, it suffers from the bias accumulated during trajectory rollouts. 
Thus, it is unable to successfully solve the tasks that require decision-making over a long horizon~\cite{mendonca2020meta}.
Furthermore, the resulting deterministic model neglects uncertainty and is known to be vulnerable to both underfitting and overfitting~\cite{chua2018deep}.
Thus, the control actions are often optimistically chosen even when conservative actions are needed to account for the unreliability of the learned models. The resulting policy fails to safely drive the robot, thereby degrading the overall performance. 
Another limitation of GrBAL is the lack of a systematic exploration scheme, as opposed to our method that leverages the exploration capability of SAC.
MPC-SAC shows slightly better results compared to GrBAL while underperforming all the other algorithms. This indicates the importance of the context encoder for efficiently learning the task distribution, as SAC alone is unable to adapt to constantly changing tasks despite the aid of MPC.
In summary, our systematic infusion of MPC actions accelerates meta-training, particularly in the early stages. However, the infusion rate should be kept under a certain threshold to improve the overall performance.

\newcolumntype{M}[1]{>{\centering\arraybackslash}m{#1}}
\begin{table} 
	\caption{Performance averaged over 24 unseen tasks. Both PEARL and MPC-PEARL are tested with and without online adaptation in a single trajectory.
    The performance without online adaptation is tested after experiencing two trajectories.}
	\setlength{\tabcolsep}{0.08cm}
	\renewcommand{\arraystretch}{1.2}
	\begin{center}
		\scalebox{1}{%
			\begin{tabular}{|c|c|M{0.9cm}|M{0.75cm}|M{0.75cm}|M{0.75cm}|M{0.75cm}|M{0.75cm}|M{0.75cm}|}
				  \hline
				\multirow{2}{*}{\textbf{Metric}} & \multirow{2}{*}{\textbf{Online}} & \multirow{2}{*}{\textbf{PEARL}} & \multirow{2}{*}{\textbf{MPC}}  & \multirow{2}{*}{\shortstack{\textbf{MPC-}\\\textbf{SAC}}}& \multicolumn{4}{c|}{\textbf{MPC-PEARL} ($\varepsilon$)}\\ 
				& & & & & $0.2$ & $0.5$ & $0.8$ & $1$ \\\hline

				\multirow{2}{*}{SR} 		
				& \textsf{X}& $0.36$ & $0.67$ & $0.17$ & $\bm{0.92}$ & $0.76$ & $0.66$ & $0.35$ \\
				& \textsf{O}& $0.25$ & $-$ &  $-$ &  $0.83$                      & $0.54$ & $0.49$ & $0.25$\\
				\hline
				
				\multirow{2}{*}{TT}			
                & \textsf{X}& $64.32$ & $59.62$ & $73.33$ & $\bm{38.61}$ & $41.85$ & $44.08$ & $66.21$ \\
                & \textsf{O}& $69.34$ & $-$  & $-$    & $51.65$                    & $54.83$ & $57.67$ & $67.75$\\
                \hline

				\multirow{2}{*}{CFR} 
                & \textsf{X}& $\bm{0.99}$ & $0.83$ & $0.69$ & $\bm{0.99}$ & $0.97$ & $0.77$ & $0.54$ \\
                & \textsf{O}& $0.89$                    & $-$  &  $-$    & $0.90$ & $0.67$ & $0.60$ & $0.58$ \\
                \hline
                
                \multirow{2}{*}{SPL}			
                & \textsf{X}& $0.31$ & $0.63$ &$0.14$ & $\bm{0.78}$ & $0.72$ & $0.61$ & $0.32$\\
                & \textsf{O}& $0.20$ & $-$ & $-$   & $0.65$                    & $0.43$ & $0.37$ & $0.23$\\
                \hline
                
                \multirow{2}{*}{MGD}		
                & \textsf{X}& $33.59$ & $30.13$ & $74.90$ & $\bm{22.16}$ & $24.33$ & $22.25$                    & $22.37$ \\
                & \textsf{O}& $52.87$ & $-$ & $-$    & $45.79$ & $48.37$ & $49.92$ & $42.14$\\
                \hline
              
               \hline
		\end{tabular}
		}
		\vspace{-0.2in}
		\label{tab:results}
	\end{center}
\end{table}

Table~\ref{tab:results} shows the quantitative navigation performance of the trained MPC-PEARL policy compared to its baselines, PEARL, MPC, and MPC-SAC.
MPC-PEARL outperforms its baselines in terms of SPL for $\varepsilon=0.2, 0.5$ and MGD for all values of $\varepsilon$  in addition to SR and TT, as shown in Fig.~\ref{fig:results} (b) and (c). 
This indicates that our method effectively exploits both MPC and PEARL to attain better navigation performances.
As PEARL behaves in a conservative manner, 
it achieves the highest value of 
the safety-critical metric CFR  among all algorithms; nevertheless, MPC-PEARL with $\varepsilon = 0.2$ yields a comparable value.
Increasing $\varepsilon$ reduces the CFR, as a large $\varepsilon$ causes MPC-PEARL to rely on the MPC module to handle dynamic obstacles instead of learning how to avoid collisions.

Table~\ref{tab:results} also presents the performance tested with online adaptation in a single trajectory. 
The frequency parameter $t_\mathrm{update}$ in Algorithm~\ref{alg:PEARL_MPC_test} was chosen as $t_\mathrm{update}=5$, which is equivalent to updating the posterior $15$ times within a single trajectory. 
Overall, our online adaptation scheme does not significantly degrade the performance of the MPC-PEARL policy with $\varepsilon =0.2$
although the online version uses much less information, experiencing a single trajectory instead of two. 
These results indicate that our method is capable of handling the scenarios where adaptation is required in a single run, without the need to experience multiple test cases.

\begin{figure}[t]
\centering
\includegraphics[width=0.95\linewidth]{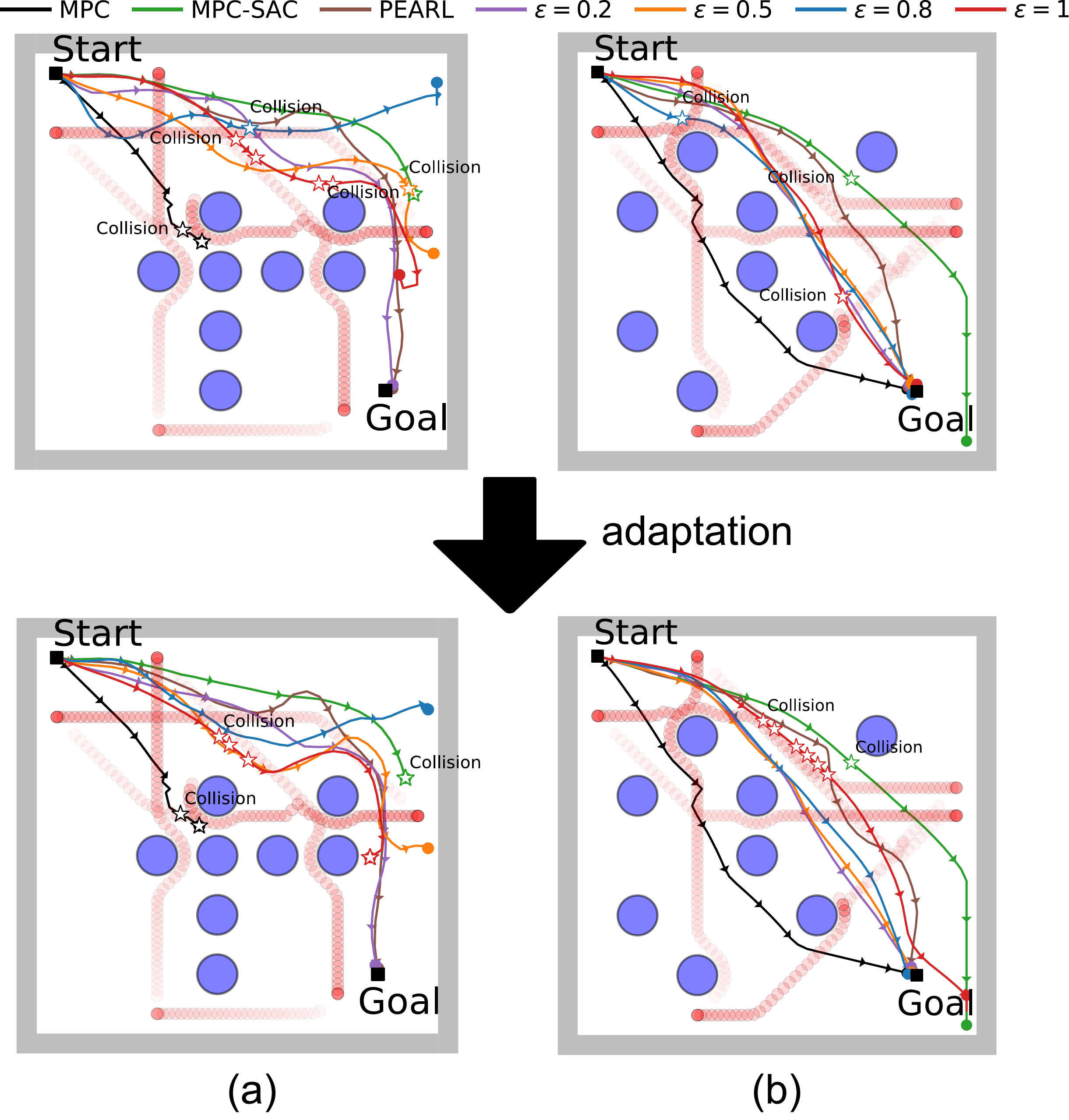}
\caption{Trajectories  before and after adaptation over 10 episodes in two  test tasks, (a) and (b). 
Dynamic and static obstacles are shown in red and blue, respectively. The location at which a collision occurs is marked with a star. }
\label{fig:trajs}
\vspace{-0.2in}
\end{figure}

Fig.~\ref{fig:trajs} shows the robot's trajectories for various controllers in two different test tasks. 
The limitation of MPC is clearly illustrated in Fig.~\ref{fig:trajs} (a), where the robot controlled by MPC is stuck in the middle due to its short horizon. 
However, MPC-PEARL with $\varepsilon = 0.2$ produces a desirable behavior, which is the result of taking into account the full horizon.
In Fig.~\ref{fig:trajs} (b), PEARL steers the robot toward a conservative path, detouring the cluttered region. In contrast, MPC does not circumvent the crowded region, achieving a more time-efficient trajectory.
In the case of MPC-PEARL with $\varepsilon =  0.2, 0.5, 0.8$, the robot takes a riskier yet shorter path without a collision, which validates the effectiveness of MPC-PEARL in guiding the robot to take near-optimal trajectories. Lastly, we confirm the inferior performance of MPC-SAC due to its poor adaptation capability to unseen tasks.

\subsection{Sidewalk Environment}

We now consider a sidewalk environment, where the movements of real-world pedestrians are generated using the UCY dataset~\cite{lerner2007crowds}.
To generate diverse scenarios from the dataset, a 6-minute video of the dataset was split  into 31 episodes of 10 seconds each.
For this environment, $N_{\mathrm{task}} = 25$, $N_{\mathrm{train}}=750$, and $T = 250$ are used. The results are tested on 6 unseen tasks using the online adaptation scheme.

\begin{figure}[t]
\centering
\includegraphics[width=0.8\linewidth]{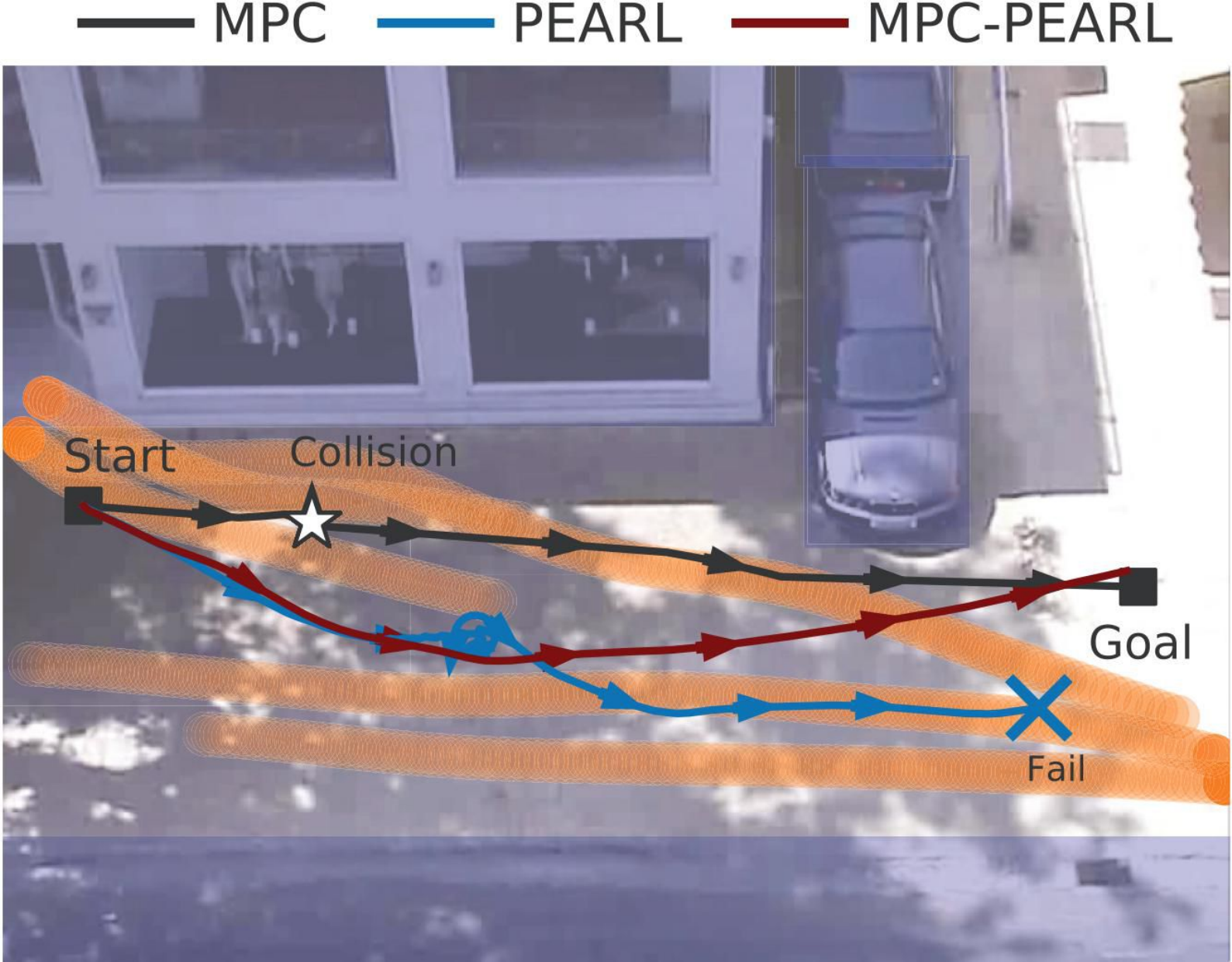}
\caption{Trajectories after online adaptation. The orange circles represent the trajectories of the pedestrians, while the static obstacles are marked by rectangles. The location at which a collision occurs is marked with a star.}
\label{fig:ucy}
\vspace{-0.1in}
\end{figure}

Fig.~\ref{fig:ucy} shows the robot trajectories obtained by MPC-PEARL with  $\varepsilon=0.2$ and using the two baselines. 
The aggressive goal-oriented nature of GP-MPC 
 causes a collision.
Specifically, GP-MPC drives the robot to the target when there is no pedestrian within its prediction horizon but fails to stay safe
when an unexpected pedestrian enters its visibility area, resulting in an SR of $0.48$ and a TT of $59.62$.
 Meanwhile, PEARL displays a conservative behavior that does not lead to any collisions, with a CFR of $0.84$, but falling behind GP-MPC in terms of other metrics, such as an SR of $0.03$ and an SPL of $0.03$. PEARL is the slowest to reach the target, with a TT of $249.61$. 
 Finally, the controller learned by MPC-PEARL successfully accomplished   the tasks with an SR of $0.67$, balancing safety and efficiency. Overall, MPC-PEARL with $\varepsilon=0.2$  showed the best performance, with a TT of $211.35$ and an SPL of $0.44$.

\section{Conclusion}\label{sec:conc}

In this work, we proposed a novel combination of meta-RL and MPC with event-triggered probabilistic switching between the two modules. 
Our method benefits from both of these complementary techniques in terms of learning stability and computational efficiency.
The results of our experiments demonstrate the superior performance of our algorithm compared to several baselines.

The proposed method can be extended in several interesting ways.
First, the latent context variable space may be expanded to also accommodate the MPC module.
Second, the method can be generalized to handle partially observable environments.
Third, a distributionally robust version of MPC can be adopted to address learning errors.

\appendix

\section*{Details of GPR}\label{app:GPR}

In each time stage, 
we use previous $N_{\mathrm{GP}}$ states and velocities of the obstacles $\tilde{\mathrm{x}} = \{x_{t-l}^d\}_{l=1}^{N_{\mathrm{GP}}}$ and $\tilde{\mathrm{v}} = \{v_{t-l}^d\}_{l=1}^{N_{\mathrm{GP}}}$, where $v_{t}^d = x_{t+1}^d - x_{t}^d$.
Given the dataset $D_t = \{\tilde{\mathrm{x}}, \tilde{\mathrm{v}}\}$, a GPR is performed to find a function $\bold{v}$ mapping the state to the corresponding velocity, i.e., 
\[
\tilde{\mathrm{v}} = \bold{v}(\tilde{\mathrm{x}}) + w,
\]
where $w$ is a zero-mean Gaussian noise with covariance $\Sigma^w = \mathrm{diag}[\sigma_{w,n_d}^2, \dots, \sigma_{w,n_d}^2]$.
The prior of the $j$th entry of $\bold{v}$ is a zero-mean Gaussian with a radial basis function kernel.
 
The corresponding approximation of $\bold{v}$ 
at an arbitrary test point $\mathbf{x}$
is then  represented by $\bold{v}(\mathbf{x}) \sim\mathcal{N}(\mu^v(\mathbf{x}), \Sigma^v(\mathbf{x}))$, where $\mu^v$ and $\Sigma^v$ are the mean and covariance of $\bold{v}$, computed as
\begin{equation}\nonumber
\begin{split}
&\mu^v_j(\mathbf{x})= K_j(\mathbf{x},\tilde{\mathrm{x}})[K_j(\tilde{\mathrm{x}},\tilde{\mathrm{x}})+\sigma_{w,j}^2I]^{-1}\tilde{\mathrm{v}}_j\\
&\Sigma^v_j(\mathbf{x}) = k_j(\mathbf{x},\mathbf{x})-K_j(\mathbf{x},\tilde{\mathrm{x}})[K_j(\tilde{\mathrm{x}},\tilde{\mathrm{x}})+\sigma_{w,j}^2I]^{-1}K_j(\tilde{\mathrm{x}},\mathbf{x}).
\end{split}
\end{equation}
Here, $K_j(\tilde{\mathrm{x}},\tilde{\mathrm{x}})$ denotes the $N_{\mathrm{GP}}$ by $N_{\mathrm{GP}}$ covariance matrix of training input data, i.e., $K_j^{(l,k)}(\tilde{\mathrm{x}},\tilde{\mathrm{x}})=k_j(\tilde{\mathrm{x}}^{(l)},\tilde{\mathrm{x}}^{(k)})$.
The resulting GP approximation $\hat{x}^d_{t+k+1}\sim \mathcal{N}(\mu_{t+k+1},\Sigma_{t+k+1})$ is obtained by propagating the state vector starting from $\hat{x}^d_{t}\sim\mathcal{N}(x^d_{t},\mathbf{0})$ and using the estimated velocity evaluated at the current mean state:
\[
\begin{split}
\mu_{t+k+1} &= \mu_{t+k} + \mu^v(\mu_{t+k}), \;\; 
\Sigma_{t+k+1} = \Sigma_{t+k} + \Sigma^v(\mu_{t+k}),
\end{split}
\]
where the input and output are assumed to be uncorrelated. 
The prediction approach using GPR is appealing when a dataset is not available in the initial time stages but is collected and updated online.

\bibliographystyle{IEEEtran}
\bibliography{reference} 

\end{document}